\pdfoutput=1
\documentclass{article}
\usepackage{spconf,amsmath,graphicx}
\usepackage{epsfig}
\usepackage{subfigure}
\usepackage{calc}
\usepackage{amssymb}
\usepackage{amstext}
\usepackage{amsthm}
\usepackage{multicol}
\usepackage{pslatex}
\usepackage[small]{caption}
\usepackage{multirow}
\usepackage{color}
\usepackage[ruled,linesnumbered]{algorithm2e}
\usepackage{algpseudocode}

\usepackage[hyphens]{url}

\DeclareMathOperator*{\argmin}{argmin}

\title{IMAGE RESTORATION AND RECONSTRUCTION USING VARIABLE SPLITTING \\
AND CLASS-ADAPTED IMAGE PRIORS}
\name{Afonso M. Teodoro \quad Jos\'{e} M. Bioucas-Dias \quad M\'{a}rio A. T. Figueiredo\thanks{This work was partially supported by the {\it Funda\c{c}\~ao para a Ci\^encia e Tecnologia} (FCT), grants UID/EEA/5008/2013 and BD/102715/2014.}}
\address{Instituto de Telecomunica\c{c}\~{o}es\\ Instituto Superior T\'{e}cnico, Universidade de Lisboa, Portugal }
\begin{document}
\ninept
\maketitle
\begin{abstract}
This paper proposes using a Gaussian mixture model as a patch-based prior, for solving two image inverse problems, namely image deblurring and compressive imaging. We capitalize on the fact that variable splitting algorithms, like ADMM, are able to decouple the handling of the observation operator from that of the regularizer, and plug a state-of-the-art algorithm into the  denoising step. Furthermore, we show that, when applied to a specific type of image, a Gaussian mixture model trained from an database of images of the same type is able to outperform current state-of-the-art generic methods.
\end{abstract}

\begin{keywords}
Variable splitting, ADMM, Gaussian mixtures, plug-and-play, image reconstruction, image restoration.
\end{keywords}
\section{Introduction}
\label{sec:intro}

The classical linear inverse problem formulation of image reconstruction or restoration has the form
\begin{equation}
{\bf y} = {\bf Ax} + \textbf{n}, \label{eq:invprob}
\end{equation}
where $\textbf{y} \in \mathbb{R}^M$ denotes the observed data, $\textbf{x} \in \mathbb{R}^N$ is the (vectorized) underlying image to be estimated, $\textbf{A}$ is the observation matrix, and $\textbf{n}$ is noise (herein assumed to be Gaussian, with zero mean and known variance $\sigma^2$). Typically, these problems are ill-posed, {\it i.e.}, there is no solution, or the solution is not unique, because ${\bf A}$ is not invertible or is extremely ill-conditioned \cite{Bertero}. Consequently, these problems can only be solved satisfactorily if a regularizing term is introduced. This function, denoted $\phi$, is used to promote characteristics that the original image is known or assumed to have. A classical approach to tackling  \eqref{eq:invprob} with the help of $\phi$ is by formulating an optimization problem of the form 
\begin{equation}
\hat{\textbf{x}} = \underset{\textbf{x}}{\argmin} \, \, \frac{1}{2} \Vert \textbf{Ax} - \textbf{y} \Vert_2^2 + \alpha\, \phi(\textbf{x}), \label{eq:analysis}
\end{equation}
which combines a data-fidelity term with the regularizer, with parameter $\alpha$ controlling their trade-off. To make problem \eqref{eq:analysis} tractable, most of the work in this area has been focused on designing convex regularizers, such as the total-variation norm \cite{rudin}, which promotes piece-wise smoothness while maintaining sharp edges, or sparsity-inducing norms on wavelet transforms or other representations \cite{Elad_Figueiredo_Ma}.

Although many optimization methods have been developed to address such problems, a lot of attention has been recently focused on variable splitting methods, such as the \textit{alternating direction method of multipliers} (ADMM). Having its roots in the 1970's \cite{gabay}, ADMM is a flexible and efficient tool, currently widely used to address imaging inverse problems \cite{FigueiredoBioucas2012}, as well as in machine learning, and other areas \cite{boyd}. ADMM  is able to deal with non-smooth convex regularization terms, it has good convergence properties, and it can be very efficiently implemented in a  distributed way \cite{boyd}.

As mentioned above, the  majority of work on imaging inverse problems has focused on convex regularizers, due to their tractability. However, it is widely accepted today that the state-of-the-art methods for image denoising ({\it i.e.}, problems of the form \eqref{eq:invprob} with $\textbf{A} = {\bf I}$) do not correspond to solving problems of the form \eqref{eq:analysis} with a convex regularizer. In fact, most (if not all) of the best performing denoisers are patch-based (as pioneered in \cite{Buades}) and use estimation tools such as collaborative filtering \cite{dabov}, Gaussian mixture models (GMM) \cite{yu}, \cite{Teodoro2015}, or learned dictionaries \cite{aharon}. Recently, the integration of state-of-the-art denoisers into ADMM has been proposed \cite{Venkatakrishnan} (under the designation ``plug-and-play"),  exploiting the ability of ADMM to decouple the handling of the observation operator from that of the regularizer/denoiser.

In this paper, we extend the ``plug-and-play" approach in the following ways. Whereas \cite{Venkatakrishnan} uses fixed denoisers, in this paper we use a GMM-based method \cite{Teodoro2015}, which opens the door to learning class-adapted models ({\it e.g.}, for faces, text, fingerprints, or specific types of medical images). Although the idea of training denoisers for specific image classes has been very recently proposed \cite{luo}, that work considers only pure denoising problems. In this paper, we show that by using a GMM-based denoiser plugged into the ADMM algorithm, we achieve state-of-the-art results in compressive image reconstruction \cite{som}, as well as in deblurring face and text images. 

This paper is organized as follows. Section 2 reviews the use of ADMM for deblurring and compressive imaging. Section 3 presents the proposed GMM-based approach. Section 4 reports experimental results, and some concluding remarks and directions for future developments are presented in Section 5.

\section{ADMM for Imaging Inverse Problems}
\label{sec:method}
As the name suggests, splitting methods proceed by splitting the objective function and dealing with each term separately, yielding simpler optimization problems.  It is straightforward to rewrite \eqref{eq:analysis} as
\begin{equation}
\hat{\textbf{x}} = \underset{\textbf{x}}{\argmin} \, \, f_1(\textbf{x}) + f_2(\textbf{x}), \label{eq:split}
\end{equation}
where $f_1(\textbf{x}) = \frac{1}{2} \Vert \textbf{Ax} - \textbf{y} \Vert_2^2$, and $f_2(\textbf{x}) = \alpha\, \phi (\textbf{x})$. Introducing a new variable $\textbf{v}$, such that $\textbf{x} = \textbf{v}$, the unconstrained problem \eqref{eq:split} can be rewritten  as a constrained one:
\begin{equation}
\hat{\textbf{x}}, \hat{\textbf{v}}  = \underset{\textbf{x,v}}{\argmin} \, \, f_1(\textbf{x}) + f_2(\textbf{v}), \hspace{0.5cm}\mbox{subject to} \quad \textbf{x} = \textbf{v}. \label{eq:const}
\end{equation}
The purpose of this splitting is that the constrained optimization problem may be easier to solve than the original one, namely via the augmented Lagrangian method (ALM), also known as the method of multipliers \cite{Hestenes}, \cite{Powell}. The ALM proceeds by alternating between the minimization of the so-called augmented Lagrangian function 
\begin{align}
\hat{\textbf{x}}, \hat{\textbf{v}} & = \underset{\textbf{x,v}}{\argmin} \, \, f_1(\textbf{x}) + f_2(\textbf{v}) + {\bf d}^T(\textbf{x} - \textbf{v}) + \frac{\mu}{2} \Vert \textbf{x} - \textbf{v}\Vert_2^2, \label{eq:unconstlong}
\end{align}
and updating the Lagrange multipliers ${\bf d}$ (see below). In ADMM, the joint minimization in \eqref{eq:unconstlong} is replaced by separate minimizations with respect to ${\bf x}$ and ${\bf v}$, and a scaled version of the Lagrange multipliers is used. The resulting algorithm finally takes the form
\begin{align}
\textbf{x}^{k+1} & := \underset{\textbf{x}}{\argmin} \, \, f_1(\textbf{x}) + \frac{\mu}{2} \Vert \textbf{x} - \textbf{v}^k - \textbf{d}^k \Vert_2^2, \label{eq:x}\\
\textbf{v}^{k+1} & := \underset{\textbf{v}}{\argmin} \, \, f_2(\textbf{v}) + \frac{\mu}{2} \Vert \textbf{x}^{k+1} - \textbf{v} - \textbf{d}^k \Vert_2^2, \label{eq:v}\\
\textbf{d}^{k+1} & := \textbf{d}^k - (\textbf{x}^{k+1} - \textbf{v}^{k+1}).
\end{align}
Notice that \eqref{eq:x} and \eqref{eq:v} are, by definition, the {\it Moreau proximity operators} (MPO, see \cite{Bauschke}) of $f_1$ and $f_2$, computed at $\textbf{v}^k + \textbf{d}^k$ and $\textbf{x}^{k+1} - \textbf{d}^k$, respectively. Recall that the MPO of some convex function $\psi$, computed at some point ${\bf z}$, is defined as
\begin{equation}
\mbox{prox}_{\psi}(\textbf{z}) = \underset{\textbf{x}}{\argmin} \, \, \frac{1}{2} \Vert \textbf{x} - \textbf{z} \Vert_2^2 +  \psi(\textbf{x}), \label{eq:prox}
\end{equation}
and can be seen as the solution to a pure denoising problem, with $\psi$ as the regularizer and ${\bf z}$ the noisy observation. A famous MPO is the \textit{soft-threshold} function, which results from $\psi(\textbf{x}) = \Vert \textbf{x}\Vert_1$. 

Instantiating ADMM to problem  \eqref{eq:analysis}, yields the SALSA algorithm \cite{afonso}. In particular, \eqref{eq:x} becomes a quadratic optimization problem, which has a linear solution given by:
\begin{equation}
\textbf{x}^{k+1} = (\textbf{A}^T\textbf{A} + \mu \, \textbf{I})^{-1}(\textbf{A}^T\textbf{y} + \mu \, (\textbf{v}^k + \textbf{d}^k)). \label{eq:inv}
\end{equation}
As shown in \cite{afonso}, \cite{afonso2011}, this inversion can be done efficiently in several relevant classes of inverse problems, namely cyclic deblurring, inpainting, partial Fourier observations; more recently, it was shown that this inversion can also be efficiently done in non-cyclic deblurring \cite{Almeida2013}. In this paper, we will consider cyclic deblurring and compressive imaging. 

\subsection{Cyclic Deblurring}
In the case of cyclic/periodic deblurring, $\textbf{A} \in \mathbb{R}^{N\times N}$  is  a block circulant matrix, thus the inversion can be done in the 2D discrete Fourier domain. In fact, since $\textbf{A}$ is block circulant, it can be factored as $\textbf{A} = \textbf{U}^H\textbf{DU}$, where $\textbf{U}$ represents the 2D discrete Fourier transform matrix, $\textbf{U}^H = \textbf{U}^{-1}$ is its inverse (with $(\cdot )^H$ denoting conjugate transpose), and $\textbf{D}$ is a diagonal matrix. Consequently,
\[
(\textbf{A}^T\textbf{A} + \mu \,\textbf{I})^{-1}  = (\textbf{U}^H\textbf{DU}\textbf{U}^H\textbf{D}^H\textbf{U} + \mu \,\textbf{I})^{-1} 
= \textbf{U}^H(\vert \textbf{D} \vert^2 + \mu\, \textbf{I})^{-1}\textbf{U}.
\]
The inversion of the diagonal matrix $\vert \textbf{D} \vert^2 + \mu \textbf{I}$ has linear cost, and the multiplications by $\textbf{U}$ and $\textbf{U}^H$ can be done via the FFT algorithm.

\subsection{Compressive Imaging}
In compressive imaging \cite{Romberg}, \cite{som}, $\textbf{A} \in \mathbb{R}^{M\times N}$ is a Gaussian measurement matrix, with $M  < N$. The solution to \eqref{eq:x} is again given by the $N\times N$ inversion in \eqref{eq:inv}, which becomes the  bottleneck of the algorithm. Resorting to the matrix inversion lemma, it is possible to reduce the cost of the inversion by a factor of roughly $(N/M)^3$:
\begin{equation}
(\textbf{A}^T \textbf{A} + \mu \textbf{I})^{-1} = \frac{1}{\mu} \left( I - \textbf{A}^T ( \textbf{A A}^T + \mu \textbf{I} )^{-1} \textbf{A} \right),
\end{equation} 
where the required inversion is now of size $M \times M$. Note, also, that this inversion is done only once, and can be precomputed and stored.

\section{Plug-and-Play Priors}
In this paper, instead of using the MPO of a convex regularizer in \eqref{eq:v}, we implement that step by using a state-of-the-art denoiser, as in \cite{Venkatakrishnan}, exploiting the fact that the MPO is itself a denoising function. Whereas \cite{Venkatakrishnan} uses  BM3D \cite{dabov} and  K-SVD \cite{aharon}, we take an alternative route and adopt GMM-based denoising \cite{Teodoro2015}, \cite{ZoranWeiss}. 

As shown in  \cite{ZoranWeiss}, clean image patches are well modeled by a GMM, which can be estimated from a collection of noiseless image patches. In \cite{Teodoro2015},
we showed that this GMM can be directly estimated from the noisy image itself, using the expectation-maximization (EM) algorithm. With a GMM prior for the clean patches in hand, the corresponding minimum mean squared error (MMSE) estimator can be obtained in closed-form  (see details in \cite{Teodoro2015}). In this paper, rather than learning the GMM prior from the observed data (which may not be possible in image deblurring, and even less so in compressive imaging), we propose to learn the GMM prior from a set of clean images. However, rather than using a collection of generic natural images, we propose to learn the GMM prior from a collection of images from a specific class, making this prior adapted to that class. The goal is to achieve better performance on this class of images  than with a generic denoiser.


The fact that a denoiser that may not correspond to the MPO of a convex regularizer is used to implement \eqref{eq:v} makes the convergence of the resulting algorithm hard to analyse. In this paper, we refrain from  theoretical convergence concerns, and focus only on the empirical performance of the method.

	\begin{figure*}[!hbtp]
		\centering
		\resizebox{0.95\textwidth}{!}{
			\begin{tabular}{cccc}
				\includegraphics[width=.21\textwidth]{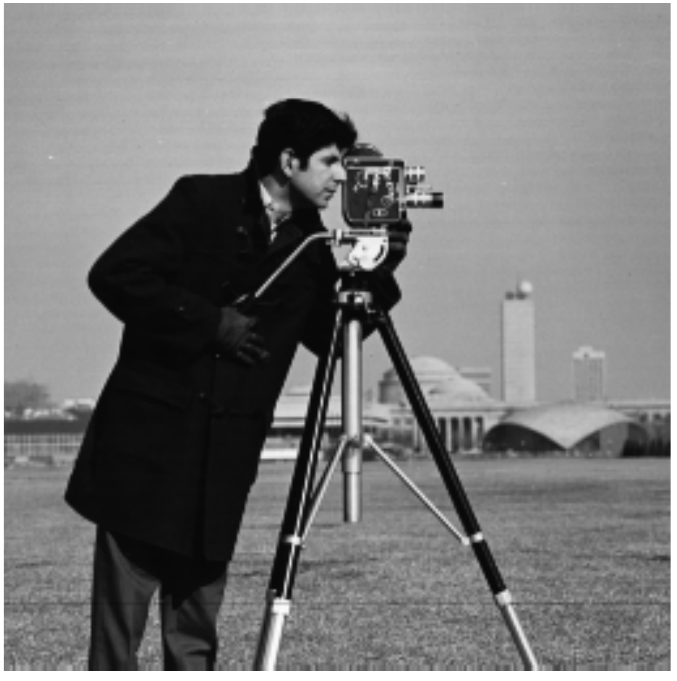}&
				\includegraphics[width=.21\textwidth]{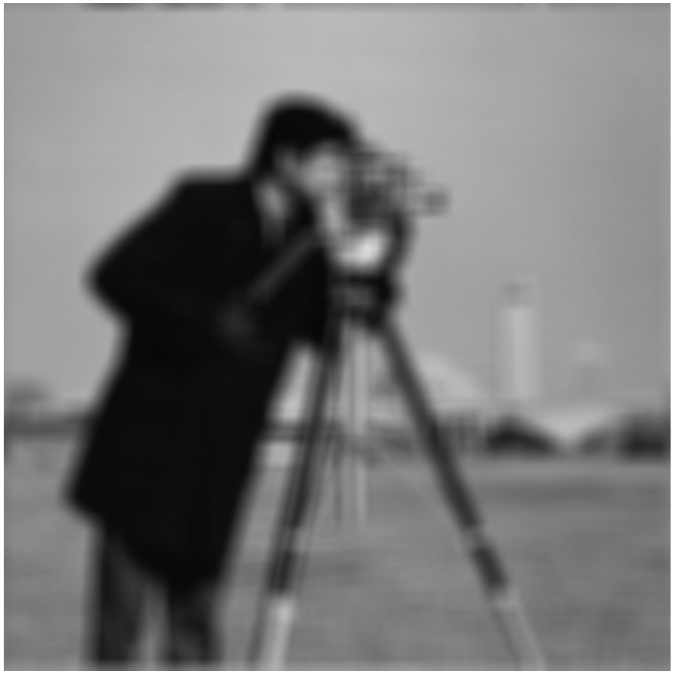}&
				\includegraphics[width=.21\textwidth]{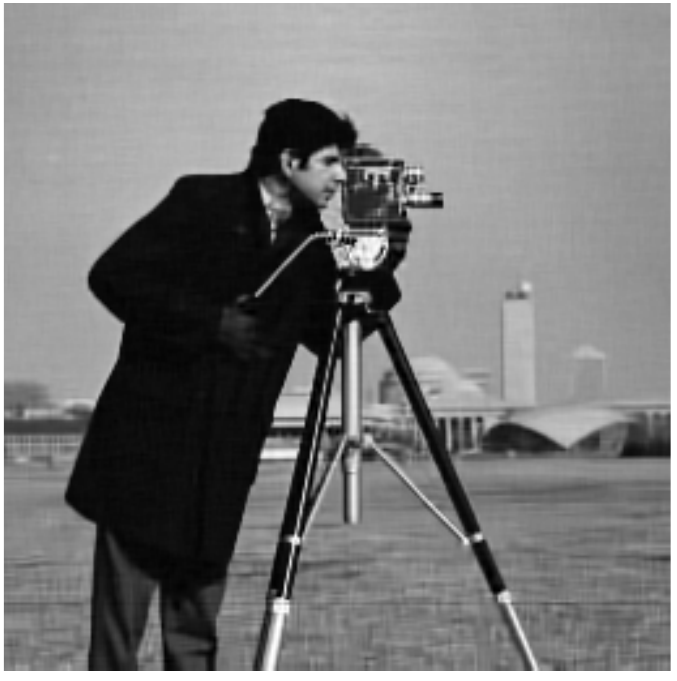}&
				\includegraphics[width=.21\textwidth]{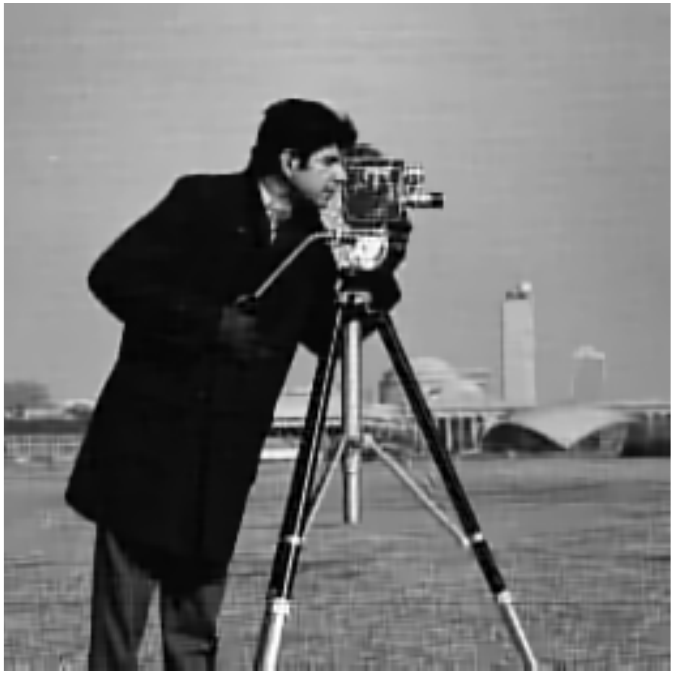}\\
				(a)&(b)&(c)&(d)
			\end{tabular}}
			\caption{Deblurring: (a) original Cameraman image; (b) blurred image (Experiment 3); (c) IDD-BM3D \cite{danielyan}; (d) ADMM-GMM. }
			\label{fig:deb1}
		\end{figure*}
		
\section{Experiments}
The proposed approach was tested with the two types of observation operators mentioned above (periodic convolution and compressive imaging). In addition to the proposed GMM denoiser, we also consider BM3D plugged into the ADMM algorithm \cite{Venkatakrishnan}, and the state-of-the-art deblurring algorithm IDD-BM3D \cite{danielyan} (with default parameters) as a benchmark. The experiments were carried out with the following four sets of images:

\vspace{0.1cm}
{\noindent \bf Generic:} this dataset comprises several benchmarks images, such \textit{Lena} and \textit{Cameraman}. The GMM-based denoiser  starts by using a mixture estimated from five other clean images (\textit{Hill}, \textit{Boat}, \textit{Couple}, \textit{Peppers}, \textit{Man});  after 100 iterations, a new GMM is obtained from the current image estimate, which aims at obtaining a GMM that is more adapted to the underlying image, and improves the final results.

\begin{table*}
\caption{ISNR on image deblurring. Methods: IDD-BM3D \cite{danielyan}; ADMM with GMM and BM3D \cite{dabov} denoising.\label{tab:deb1}}
\vspace{-0.5cm}
\begin{center}
\resizebox{0.95\textwidth}{!}{
\begin{tabular}{c|c|c|c|c|c|c|c|c|c|c|c|c}
\hline 
  {\bf Image:}         & \multicolumn{6}{c|}{Cameraman}                & \multicolumn{6}{c}{House}                    \\ \hline
{\bf Experiment:} & 1     & 2     & 3     & 4     & 5     & 6     & 1     & 2     & 3     & 4     & 5     & 6     \\ \hline
BSNR       & 31.87 & 25.85 & 40.00 & 18.53 & 29.19 & 17.76 & 29.16 & 23.14 & 40.00 & 15.99 & 26.61 & 15.15 \\ \hline
Input PSNR & 22.23 & 22.16 & 20.76 & 24.62 & 23.36 & 29.82 & 25.61 & 25.46 & 24.11 & 28.06 & 27.81 & 29.98 \\ \hline \hline
IDD-BM3D   &   \textbf{8.85}    &  \textbf{ 7.12}    &   \textbf{10.45}    &   \textbf{3.98}    &   \textbf{4.31}    &   4.89    &   \textbf{9.95}    &   \textbf{8.55 }   &   12.89    & \textbf{  5.79 }   &   \textbf{5.74}    &  7.13   \\ \hline
ADMM-GMM   &   8.34    &   6.39    &    9.73   &    3.49   &    4.18   &   \textbf{4.90}    &    9.84   &      8.40 &   12.87    &    5.57   &   5.55    &    6.65   \\ \hline
ADMM-BM3D   &   8.18    &    6.13   &    9.58   &    3.26   &    3.93  &    4.88   &   9.64    &    8.02   &   \textbf{12.95}    &    5.23   &   5.06    &    \textbf{7.37 }  \\ \hline \hline
  {\bf Image:}         & \multicolumn{6}{c|}{Lena}                     & \multicolumn{6}{c}{Barbara}                  \\ \hline
{\bf Experiment:} & 1     & 2     & 3     & 4     & 5     & 6     & 1     & 2     & 3     & 4     & 5     & 6     \\ \hline
BSNR       & 29.89 & 23.87 & 40.00 & 16.47 & 27.18 & 15.52 & 30.81 & 24.79 & 40.00 & 17.35 & 28.07 & 16.59 \\ \hline
Input PSNR & 27.25 & 27.04 & 25.84 & 28.81 & 29.16 & 30.03 & 23.34 & 23.25 & 22.49 & 24.22 & 23.77 & 29.78 \\ \hline \hline
IDD-BM3D   &    7.97   &    \textbf{6.61}   &   8.91    &   \textbf{ 4.97}   &    \textbf{4.85}   &    6.34   &    7.64   &   \textbf{3.96 }   &   \textbf{6.05 }   &   1.88    &    1.16   &   5.45    \\ \hline
ADMM-GMM   &   \textbf{8.01}    &    6.53   &   8.95    &   4.93    &      4.81 &    6.09   &   5.91    &    2.19   &   5.37    &   1.42  &    1.24   &    5.14   \\ \hline
ADMM-BM3D  &   8.00    &   6.56    &   \textbf{9.00}    &   4.88    &   4.67    &  \textbf{6.42}     &   7.32    &   2.99    &    \textbf{6.05}   &    1.55   &    \textbf{1.40}   &    5.76   \\ \hline 
\end{tabular}
}
\end{center}
\end{table*}

\vspace{0.1cm}
{\noindent \bf Text:} this dataset contains 10 images, available from the author of \cite{luo} (\url{http://videoprocessing.ucsd.edu/~eluo/}). One of them is selected as input image and all the others are used to train the GMM. Since the input image is from the same class as the images used for training, the mixture adaptation step is not performed.


\begin{figure*}
	\centering
\resizebox{0.9\textwidth}{!}{
\begin{tabular}{cccc}
\includegraphics[width=127px, height = 104px]{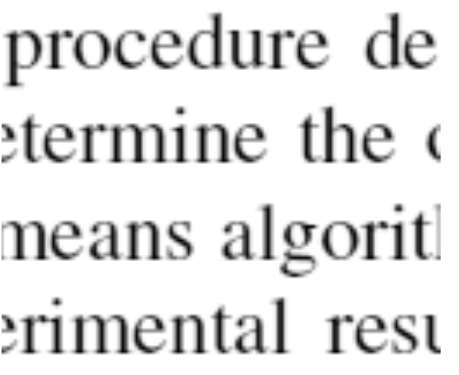}&
\includegraphics[width=127px, height = 104px]{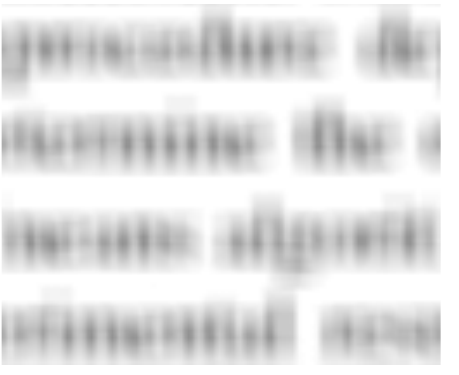}&
\includegraphics[width=127px, height = 104px]{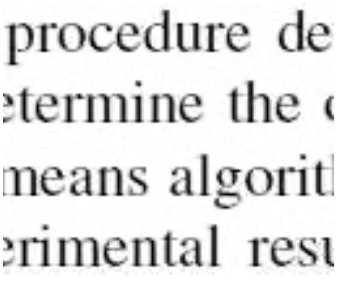}&
\includegraphics[width=127px, height = 104px]{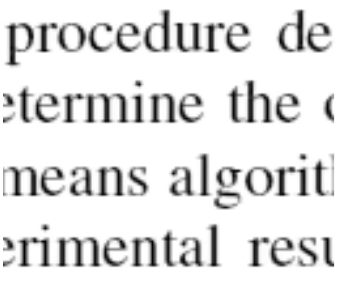}\\
(a)&(b)&(c)&(d)
\end{tabular}}
\caption{Deblurring: (a) original Text image; (b) blurred image (Experiment 3); (c) IDD-BM3D \cite{danielyan}; (d) ADMM-GMM. }
\label{fig:deb2}
\end{figure*}

\vspace{0.1cm}
{\noindent \bf Faces:} this dataset is made of 100 face images of the same subject, obtained from the same source as the text images.

\vspace{0.1cm}
{\noindent \bf Microsoft:} this dataset contains 591 images \url{http://research.microsoft.com/en-us/projects/objectclassrecognition/}. 


\begin{figure}

\resizebox{0.5\textwidth}{!}{
\begin{tabular}{cccc}
\includegraphics[width=60px, height = 80px]{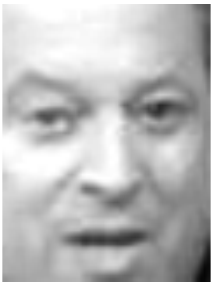}&
\includegraphics[width=60px, height = 80px]{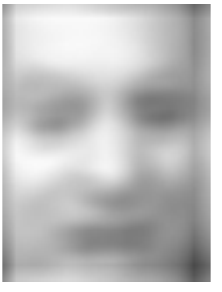}&
\includegraphics[width=60px, height = 80px]{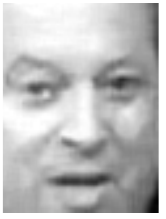}&
\includegraphics[width=60px, height = 80px]{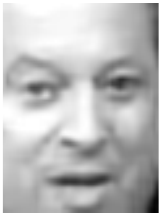}\\
(a)&(b)&(c)&(d)
\end{tabular}}
\caption{Deblurring: (a) original Face image; (b) blurred image (Experiment 3); (c) IDD-BM3D \cite{danielyan}; (d) ADMM-GMM. }
\label{fig:deb3}
\end{figure}

\begin{table*}
\caption{ISNR on image deblurring - Methods: IDD-BM3D \cite{danielyan}; ADMM with GMM prior and targeted databases.\label{tab:deb2}}
\vspace{-0.5cm}
\begin{center}
\resizebox{0.95\textwidth}{!}{
\begin{tabular}{c|c|c|c|c|c|c|c|c|c|c|c|c}
\hline 
   {\bf Image class:}        & \multicolumn{6}{c|}{Text}                & \multicolumn{6}{c}{Face}                    \\ \hline
{\bf Experiment:} & 1     & 2     & 3     & 4     & 5     & 6     & 1     & 2     & 3     & 4     & 5     & 6     \\ \hline
BSNR       & 26.07 & 20.05 & 40.00 & 15.95 & 24.78 & 18.11 & 28.28 & 22.26 & 40.00 & 15.89 & 26.22 & 15.37 \\ \hline
Input PSNR & 14.14 & 14.13 & 12.13 & 16.83 & 14.48 & 28.73 & 25.61 & 22.54 & 20.71 & 26.49 & 24.79 & 30.03 \\ \hline \hline
IDD-BM3D   &    11.97   &    8.91   &     16.29  &    5.88   &    6.81   &    4.87   &    13.66   &   11.16    &  14.96     &    7.31   &    10.33   &    6.18   \\ \hline
ADMM-GMM   &   \textbf{16.24}    &    \textbf{11.55 }  &   \textbf{23.11}    &   \textbf{8.88 }   &    \textbf{10.77}   &   \textbf{8.34 }   &   \textbf{15.05}    &   \textbf{12.59}    &    \textbf{17.28}   &  \textbf{  8.84}   &   \textbf{11.69}    &   \textbf{ 7.32}   \\ \hline 
\end{tabular}
}
\end{center}
\end{table*}

\begin{figure}[hbt]
	\centering
\resizebox{0.45\textwidth}{!}{
\begin{tabular}{cc}
\includegraphics[width=.21\textwidth]{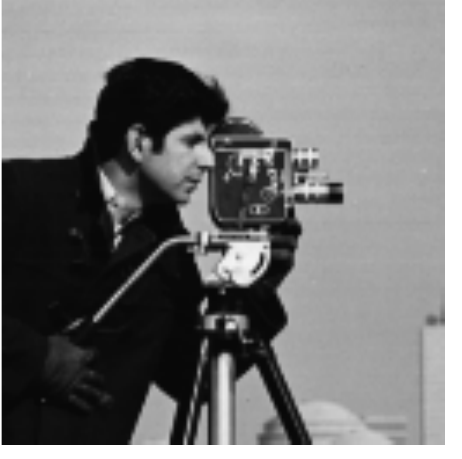}&
\includegraphics[width=.21\textwidth]{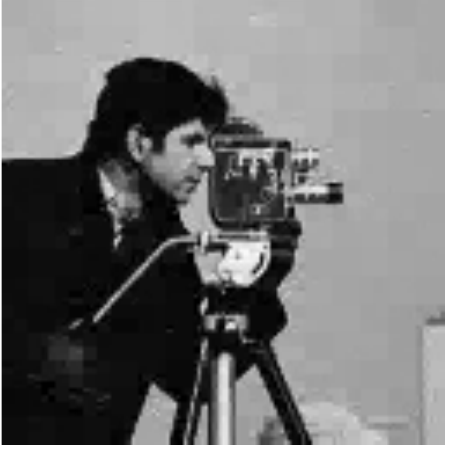} \\
(a)&(b) \\ \ \\
\includegraphics[width=.21\textwidth]{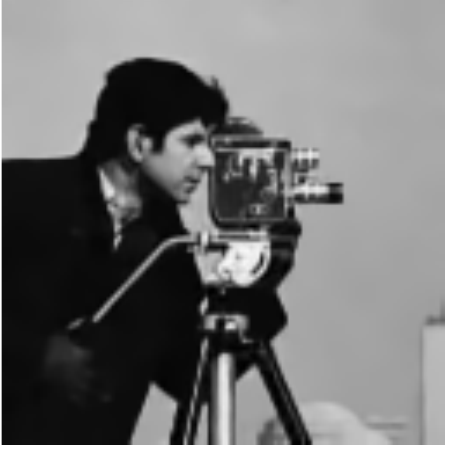} &
\includegraphics[width=.21\textwidth]{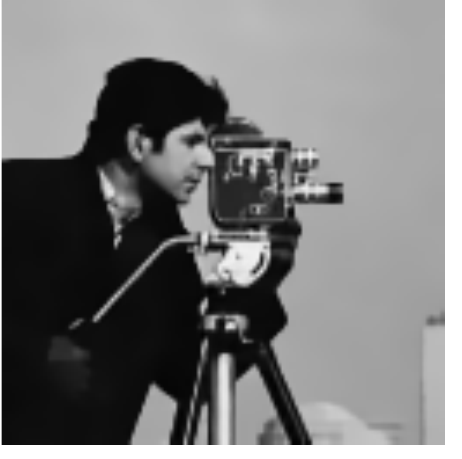}\\
(c)&(d)
\end{tabular}}
\caption{Compressive Imaging: (a) original image; (b) Turbo-GM (NMSE = $ -26.02$ dB) \cite{som}; (c) ADMM-GMM without GMM update (NMSE = $-28.41$ dB) ; (d) ADMM-GMM with GMM update (NMSE = $-30.04$ dB) }
\label{fig:cs}
\end{figure}

\subsection{Deblurring Results}
We considered the six different blur kernels available  in the BM3D package (referred to in Tables 1 and 2 as experiments 1 to 6). The results on the generic dataset are presented, both in terms of {\it improvement in SNR} (ISNR) and visual quality, in Table~\ref{tab:deb1} and in Fig.~\ref{fig:deb1}. The numbers relative to IDD-BM3D were obtained from  \cite{danielyan}, whereas the results of ADMM-BM3D were obtained by using the available implementation of BM3D in the ADMM loop, without any change. The results of the GMM prior were obtained using $6 \times 6$ patches, with 20-component mixtures. Other parameters of the algorithm, namely $\mu$, were hand tuned for the best results. These results show that the proposed approach is competitive, yet does not beat IDD-BM3D  in this
generic image deblurring experiment. Arguably, this may be due to the fact that some of the images being restored contain structures that are totally absent from the training set (e.g., the stripes on Barbara's trousers or the pattern on the table cloth).

In deblurring images of specific classes, the conclusions are remarkably different. As shown in Table~\ref{tab:deb2} and Figs.~\ref{fig:deb2} and \ref{fig:deb3}, ADMM-GMM clearly outperforms IDD-BM3D (using its default parameters) both visually and in terms of ISNR.

\subsection{Compressive Imaging Results}
To assess the performance of the proposed method in compressive imaging, the same tests as in \cite{som} were performed. Fig.~\ref{fig:cs} compares the  results (on the Cameraman image, from $M = 5000$ measurements) in terms of visual quality and normalized MSE (NMSE $ = \|{\bf x}-\hat{\bf x}\|_2^2/\|{\bf x}\|^2_2$ \cite{som}), showing that ADMM-GMM performs better than the method proposed in \cite{som}, called Turbo-GM\footnote{Note: the Turbo-GM results were obtained by running the publicly available implementation with seed 0 on the random number generator, for comparison purposes. Although the results for Turbo-GM herein presented are slightly worse than those reported in \cite{som}, the difference does not affect our conclusion: on average, ADMM-GMM performs better than Turbo-GM.}.  As in deblurring, the ADMM-GMM algorithm uses $6\times 6$ patches and a 20-component GMM. The algorithm starts with a mixture trained on images from the generic dataset; the algorithm is run for 50 iterations, with mixture update every 25 iterations. Parameter $\mu$ was kept fixed at 1. 

\begin{figure}[!hbtp]
%
\begin{minipage}[b]{.45\linewidth}
  \centering
  \centerline{\includegraphics[width=4.1cm]{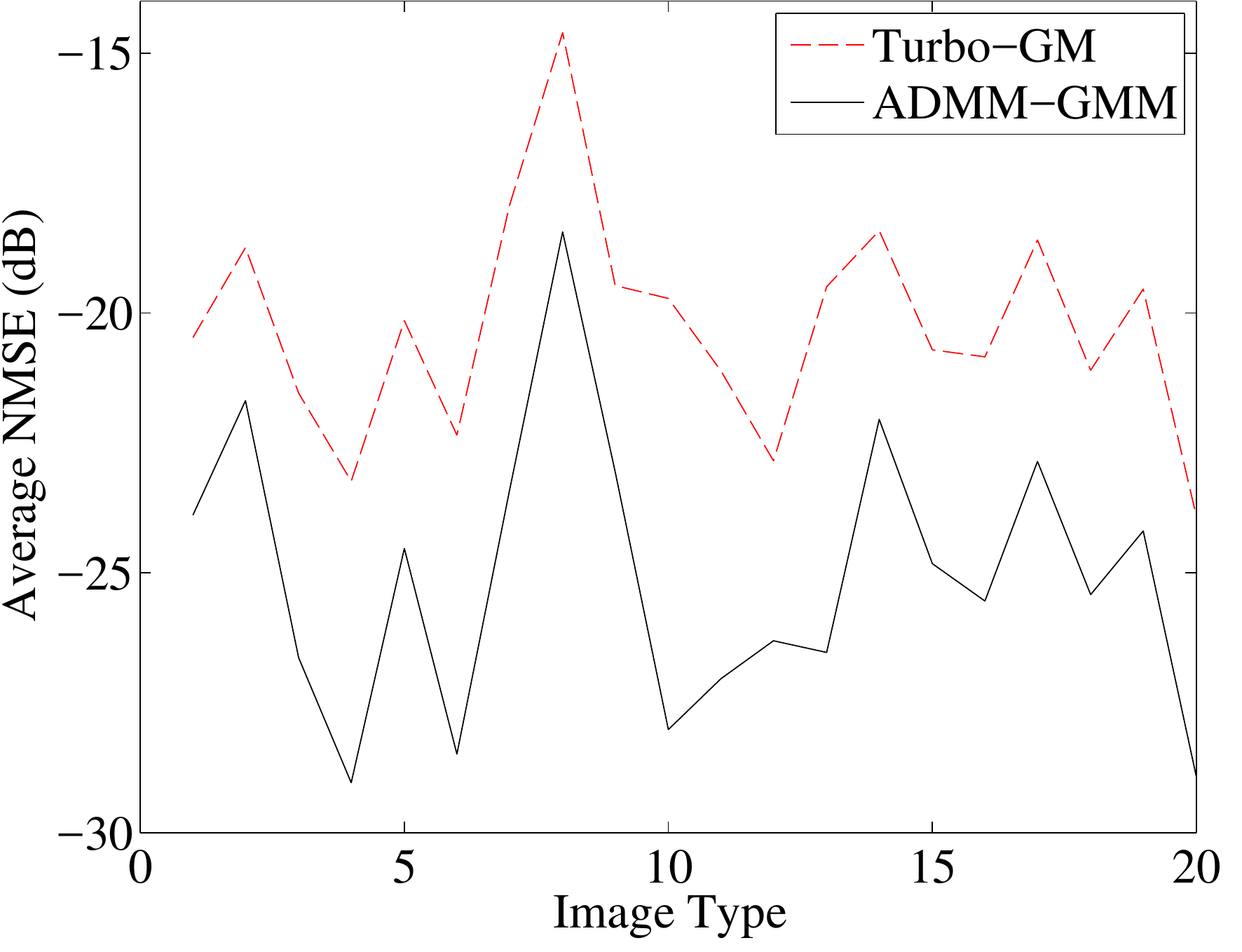}}
  \centerline{(a) NMSE for each image type.}\medskip
\end{minipage}
\hfill
\begin{minipage}[b]{0.45\linewidth}
  \centering
  \centerline{\includegraphics[width=4.1cm]{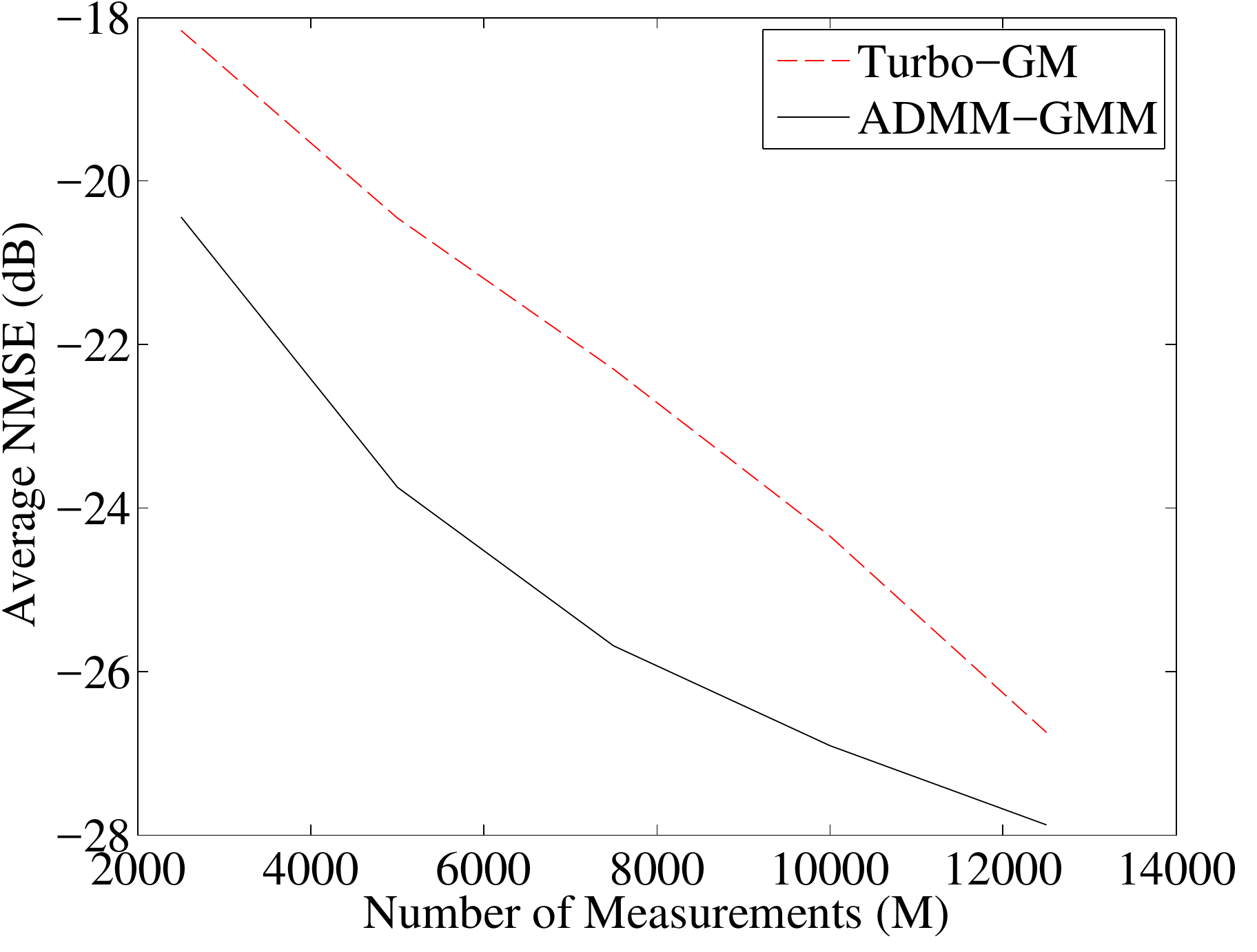}}
  \centerline{(b) NMSE for type 1 images. }\medskip
\end{minipage}
\caption{NMSE comparison on compressive imaging.}
\label{fig:csavg}
\end{figure}

Experiments were also performed on all the images in the Microsoft dataset; as in \cite{som}, each image was cropped to $192\times 192$,  from the top left corner, then resized to $128\times 128$. The results (for $M=5000$ measurements) are summarized in Fig.~\ref{fig:csavg} (a), clearly showing that for every image class, ADMM-GMM performs better than Turbo-GM, by at least 2 dB. Finally, Fig.~\ref{fig:csavg} (b) shows NMSE versus the number of measurements $M$, on type 1 images from the Microsoft dataset, showing the superiority of ADMM-GMM over Turbo-GM for a wide range of values of $M$.


\section{Conclusions and Future Work}\label{sec:conclusions}
In this paper,  we have proposed a method for class-adapted image restoration/reconstruction, by building upon the so-called \textit{plug-and-play} approach \cite{Venkatakrishnan}, and plugging class-adapted denoisers based on Gaussian mixture models (GMM) into the iterations of an ADMM algorithm. Experiments reported in this paper (both in deblurring and compressive image reconstruction) have shown that the proposed method yields state-of-the-art results when applied to  images known to contain text or a face, clearly outperforming the best generic techniques, such as IDD-BM3D \cite{danielyan}. 

Naturally,  there are still several aspects that demand further work, namely the theoretical convergence properties of plug-and-play ADMM with a GMM-based denoiser, and the optimal setting of the algorithm parameters.

\end{document}